\documentclass[10pt,twocolumn,letterpaper]{article}

\usepackage{cvpr}              % To produce the CAMERA-READY version
\usepackage{indentfirst}
\usepackage{colortbl}
\usepackage[table]{xcolor}
\usepackage{graphicx}
\usepackage{amsmath}
\usepackage{multicol}
\usepackage{multirow}
\usepackage{amsmath}
\usepackage{xcolor}
\usepackage{float}
\usepackage{stfloats}
\usepackage{placeins}
\usepackage{afterpage}
\usepackage{caption}

\definecolor{blueviolet}{RGB}{100, 20, 150}
\definecolor{darkgreen}{RGB}{0, 90, 110}

\newcommand{\bv}[1]{{\color{blueviolet} #1}} % Blue-violet color command
\newcommand{\dg}[1]{{\color{darkgreen} #1}}  % Dark green color command
\def\c\mathbf{c}
\definecolor{cvprblue}{rgb}{0.21,0.49,0.74}
\usepackage[pagebackref,breaklinks,colorlinks,allcolors=cvprblue]{hyperref}

\def\paperID{17526} % *** Enter the Paper ID here
\def\confName{CVPR}
\def\confYear{2025}

\title{RealisticDreamer: Guidance Score Distillation for Few-shot Gaussian Splatting}

\author{
Ruocheng Wu$^{1}$ \quad
Haolan He$^{1}$ \quad
Yufei Wang$^{2}$ \quad
Zhihao Li$^{2}$ \quad
Bihan Wen$^{2}$\thanks{Corresponding author.} \\
$^{1}$University of Electronic Science and Technology of China \\
$^{2}$Nanyang Technological University \\
{\tt\small wuruocheng333@outlook.com, HaolanHe7777@gmail.com}\\
{\tt\small yufei001@e.ntu.edu.sg, zhihao.li@ntu.edu.sg, bihan.wen@ntu.edu.sg}
}

\definecolor{mypink}{RGB}{255,105,180} % 较深的粉色（Hot Pink）

\begin{document}
\maketitle
\begin{abstract}
3D Gaussian Splatting (3DGS) has recently gained great attention in the 3D scene representation for its high-quality real-time rendering capabilities. However, when the input comprises sparse training views, 3DGS is prone to overfitting, primarily due to the lack of intermediate-view supervision. Inspired by the recent success of Video Diffusion Models (VDM), we propose a framework called Guidance Score Distillation (GSD) to extract the rich multi-view consistency priors from pretrained VDMs. Building on the insights from Score Distillation Sampling (SDS), GSD supervises rendered images from multiple neighboring views, guiding the Gaussian splatting representation towards the generative direction of VDM. However, the generative direction often involves object motion and random camera trajectories, making it challenging for direct supervision in the optimization process. To address this problem, we introduce an unified guidance form to correct the noise prediction result of VDM. Specifically, we incorporate both a depth warp guidance based on real depth maps and a guidance based on semantic image features, ensuring that the score update direction from VDM aligns with the correct camera pose and accurate geometry. Experimental results show that our method outperforms existing approaches across multiple datasets.
\end{abstract}    
\begin{figure}[!ht]
    \scriptsize
    \setlength{\tabcolsep}{0.7pt}
    \centering
    \includegraphics[width=0.49\textwidth]{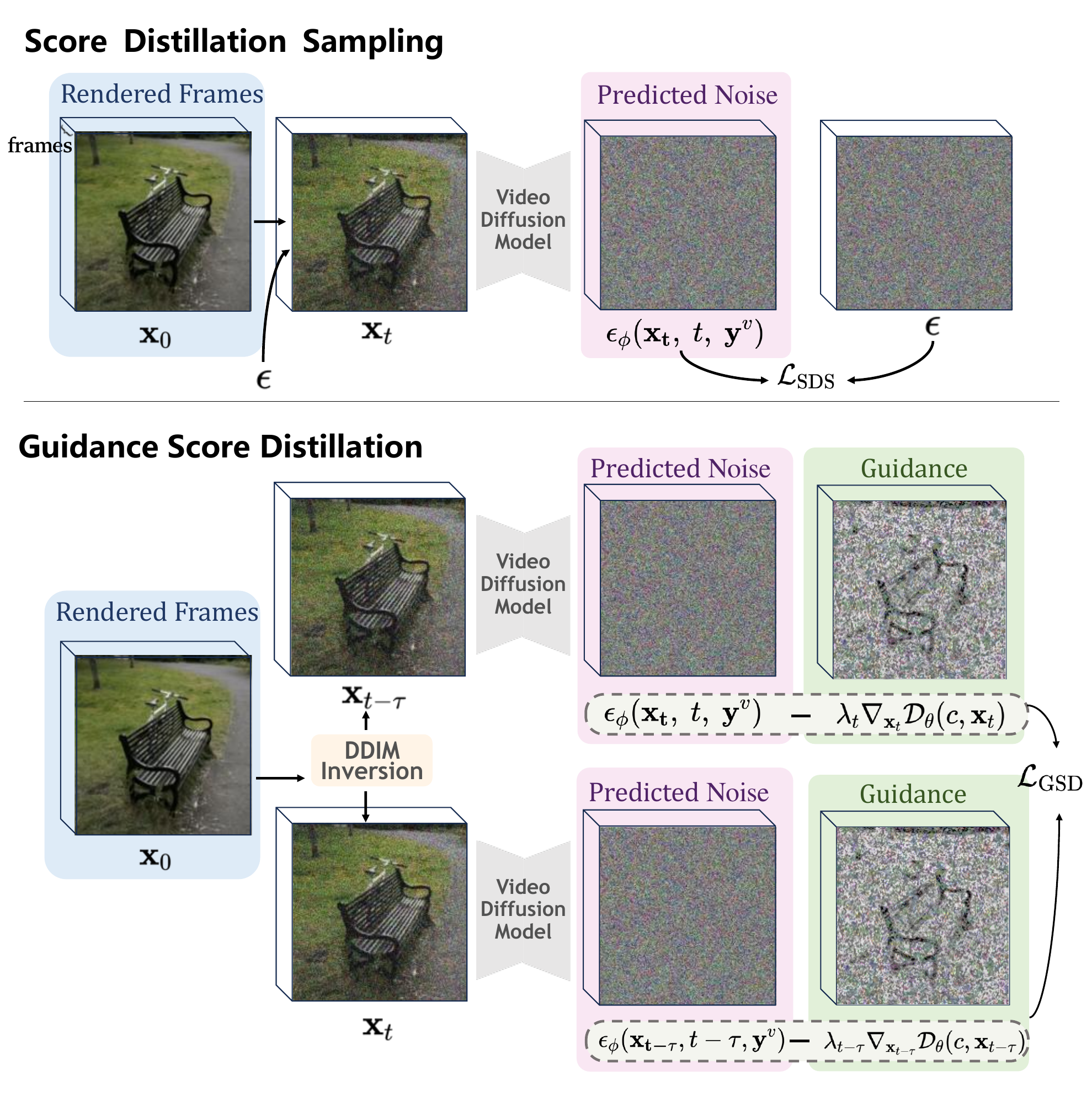}
    \caption{Comparison of Score Distillation Sampling (SDS)~\cite{poole2022dreamfusion} and the proposed Guidance Score Distillation (GSD) frameworks for video diffusion models. Our method introduces DDIM inversion~\cite{song2020denoising} while also correcting the prediction noise by guidance.} 
    \label{head_pic}
    \vspace{-3mm} 
\end{figure}

\section{Introduction}
\label{sec:intro}
% Recent advances in 3D scene reconstruction, particularly with Neural Radiance Fields (NeRF)~\cite{mildenhall2021nerf}, have demonstrated the ability to generate high-quality 3D models from a small set of 2D images. By modeling scene geometry and lighting with a deep neural network and volume rendering, NeRF produces photo-realistic images from novel viewpoints, making it highly effective for applications in virtual and augmented reality.
% Recent advances in 3D scene reconstruction, particularly with Neural Radiance Fields (NeRF)~\cite{mildenhall2021nerf}, have demonstrated the ability to generate high-quality 3D models from a small set of 2D images. By modeling scene geometry and lighting with a deep neural network and volume rendering, NeRF produces photo-realistic images from novel viewpoints, making it highly effective for applications in virtual and augmented reality.

Novel view synthesis, a fundamental problem in 3D vision, has extensive applications spanning virtual reality generation, such as VR video~\cite{cho2019novel,tseng2022pseudo} and 3D gaming~\cite{watson2022novel}, as well as real-world interactive domains like autonomous driving~\cite{tonderski2024neurad,huang2023neural} and 3D printing~\cite{voleti2025sv3d}. Since the introduction of Neural Radiance Fields (NeRF)~\cite{mildenhall2021nerf}, this field has undergone transformative changes, significantly enhancing novel view rendering quality. However, NeRF's implicit representation poses challenges in terms of speed, both in optimization and rendering. For instance, Vanilla NeRF typically requires around a week of training and approximately one second to render a single novel view. Recently, 3D Gaussian Splatting~\cite{kerbl2023gs} has emerged as a promising alternative for real-time, high-quality 3D scene rendering. By leveraging an explicit representation with 3D Gaussian ellipsoids, it dramatically improves optimization speed to just a few hours while enabling real-time rendering at high frame rates.

% Recently, 3D Gaussian Splatting (3DGS)~\cite{kerbl2023gs} has emerged as a promising alternative for real-time, high-quality 3D scene rendering. Unlike NeRF's~\cite{mildenhall2021nerf} implicit representation, 3DGS models a scene using an explicit collection of 3D Gaussian ellipsoids. These ellipsoids are used to represent the scene’s geometry, providing a more direct and efficient way to render photorealistic images. 3DGS excels in rendering speed, offering real-time performance, and has demonstrated great potential in visualizing large-scale environments.

Despite their success, both NeRF and 3DGS struggle with few-shot 3D reconstruction, where limited input images lead to inaccurate reconstructions. For NeRF, techniques like depth regularization~\cite{niemeyer2022regnerf}, frequency annealing~\cite{tancik2020fourier}, and viewpoint distortion~\cite{barron2022mip} have been introduced to mitigate overfitting and improve the quality of the synthesized views when the training set is sparse. These methods aim to regularize the model’s learning representation, encouraging smoother and more consistent scene reconstructions. Similarly, 3DGS-based methods attempt to address few-shot challenges by using Gaussian densification to add intermediate splats~\cite{kerbl2023gs}, depth supervision to preserve geometric consistency~\cite{wang2023sparsenerf}, or floaters removal to eliminate non-representative Gaussians~\cite{zhu2025fsgs}.
However, these methods fall short in addressing the core limitation of few-shot view synthesis: insufficient views lead to inaccurate 3D structures and overly smooth textures. 3DGS, in particular, is susceptible to misrepresented Gaussians and floaters. To mitigate these issues, current solutions attempt  by involving additional prior from diffusion models, \textit{e.g.}, by using pseudo-view synthesis~\cite{zhu2025fsgs} or distilling 2D diffusion priors~\cite{xiong2023sparsegs}. While improved results are achieved by integrating the image prior from 2D diffusion models, they still suffers from the multi-view consistency priors, resulting in view inconsistency and artifacts.
% Existing 2D diffusion methods show promise but lack multi-view consistency priors, leading to issues with viewpoint inconsistency and inaccurate camera poses. 
% Furthermore, SDS[] itself provides only rough supervision, as the distillation process lacks direct constraints on the multi-view coherence of the scene. This means that while these methods generate plausible intermediate views, they fail to impose strong multi-view consistency, which is critical for high-quality 3D reconstruction. 

%
Inspired by the success of Video Diffusion Models (VDM)~\cite{blattmann2023stable, ho2022imagen, guo2023animatediff} to generate high-quality, multi-view consistent videos, we propose a novel framework that leverage VDM’s ability to enhance in few-shot 3D reconstruction for both viewpoint consistency and reconstruction quality without fine-tuning the whole model.
Specifically, we introduce Guidance Score Distillation (GSD) framework, which is based on the existing experience with score distillation techniques~\cite{poole2022dreamfusion,zhu2023hifa,chung2023luciddreamer,wang2023steindreamer}. Our approach leverages the existing camera views and training images as guidance to correct the predicted scores at different time steps of DDIM inversion~\cite{song2020denoising}. This method solves the issue of misalignment in position and geometry estimation for blurry images of VDMs, which typically causes incorrect score matching directions.
To be specific, we introduce two novel guidance techniques: Depth Warp Guidance and Semantic Feature Guidance, designed to enhance geometric accuracy and semantic consistency. Depth Warp Guidance is a technique inspired by recent work in depth warping~\cite{masoumian2022monocular, zhan2018unsupervised}. By leveraging established depth estimation methods, this guidance enables the warping of images from training views to novel viewpoints.In addition to depth guidance, we introduce Semantic Feature Guidance using DINO-based features~\cite{caron2021emerging}.

% effectively augmenting the available data for sparse-view scenarios. The core idea is to propagate depth information from reference views to unseen views through camera pose transformations, ensuring that the geometry remains consistent even when only sparse training views are available. This approach helps combat overfitting by enforcing 3D consistency across multiple viewpoints, particularly when the input data is limited. The image re-projection loss ensures that the warped images are aligned with the target views, guiding the model towards a more accurate and robust scene reconstruction.

 % DINO~\cite{caron2021emerging}, a self-supervised feature representation model, provides semantic cues that help the reconstruction process retain important scene details even when the input views are sparse. By aligning the feature maps from the training views with the features generated by the model during reconstruction, the model is encouraged to preserve the semantic structure of the scene. This guidance improves the overall quality of the reconstruction, particularly in terms of semantic consistency, which is crucial for handling sparse input data effectively.

% Together, Depth Warp Guidance and Feature Guidance provide complementary cues for improving few-shot 3D reconstruction. These techniques lay the foundation for our framework, which combines GSD and the VDM prior to achieve improved performance in few-shot 3D scene reconstruction. 

Our experimental results demonstrate the effectiveness of the proposed framework, showcasing significant improvements in 3D reconstruction quality compared to existing state-of-the-art methods. Overall, our contributions can be summarized as follows.

\begin{itemize}

% \item \zhihao{
% To the best of our knowledge, we are the first to introduce a video-based generative prior to tackle the information loss caused by missing viewpoints in the few-shot 3D Gaussian Splatting (3DGS) . Our proposed Guidance Score Distillation (GSD) framework seamlessly integrates the guidance capabilities of Video Diffusion Models (VDMs) with 3DGS reconstruction without fine-tuning.
% }

\item We propose Guidance Score Distillation (GSD), the framework to leverage pretrained Video Diffusion Models (VDMs) as generative priors for supervising few-shot 3D Gaussian Splatting (3DGS) w/o further fine-tuning. GSD introduces a unified guidance formulation for VDMs, steering 3DGS optimization along a corrected generative path to enhance few-shot performance.

% \item We propose the Guidance Score Distillation (GSD) framework, which extracts view-consistency priors from Video Diffusion Models (VDM) to supervise 3D scene reconstruction without fine-tuning. The GSD framework introduce a unified guidance formulation for VDM, guiding the optimization direction of 3DGS towards corrected VDM’s generative direction via multiple guidance, thereby enhancing performance under few-shot settings.

\item To ensure the generative direction of VDM is applicable to 3D reconstruction, we propose two guidance mechanisms to adjust the noise prediction direction. Specifically, we design Depth Warping Guidance based on depth prior from the pretrained the monocular estimator and Semantic Feature Guidance based on DINO features, which ensure that VDM’s score update direction aligns with accurate camera poses and geometric structure.

\item 
Experimental results show that our method outperforms existing approaches across various training view counts and datasets. Unlike prior methods such as FSGS, which suffer performance degradation compared to vanilla 3DGS as the number of views increases (\textit{e.g.}, a $\sim0.1$ dB drop with 9 views), our framework consistently enhances rendering quality, achieving a $\sim0.4$ dB improvement.

% \item Experimental results demonstrate that our method outperforms existing methods across multiple datasets, significantly improving view consistency and geometric accuracy while overcoming the inherent limitations of traditional Score Distillation Sampling (SDS) methods.
\end{itemize}
\section{Related Work}
\label{sec:formatting}

%-------------------------------------------------------------------------

\begin{figure*}[t]
\centering
\includegraphics[width=\textwidth]{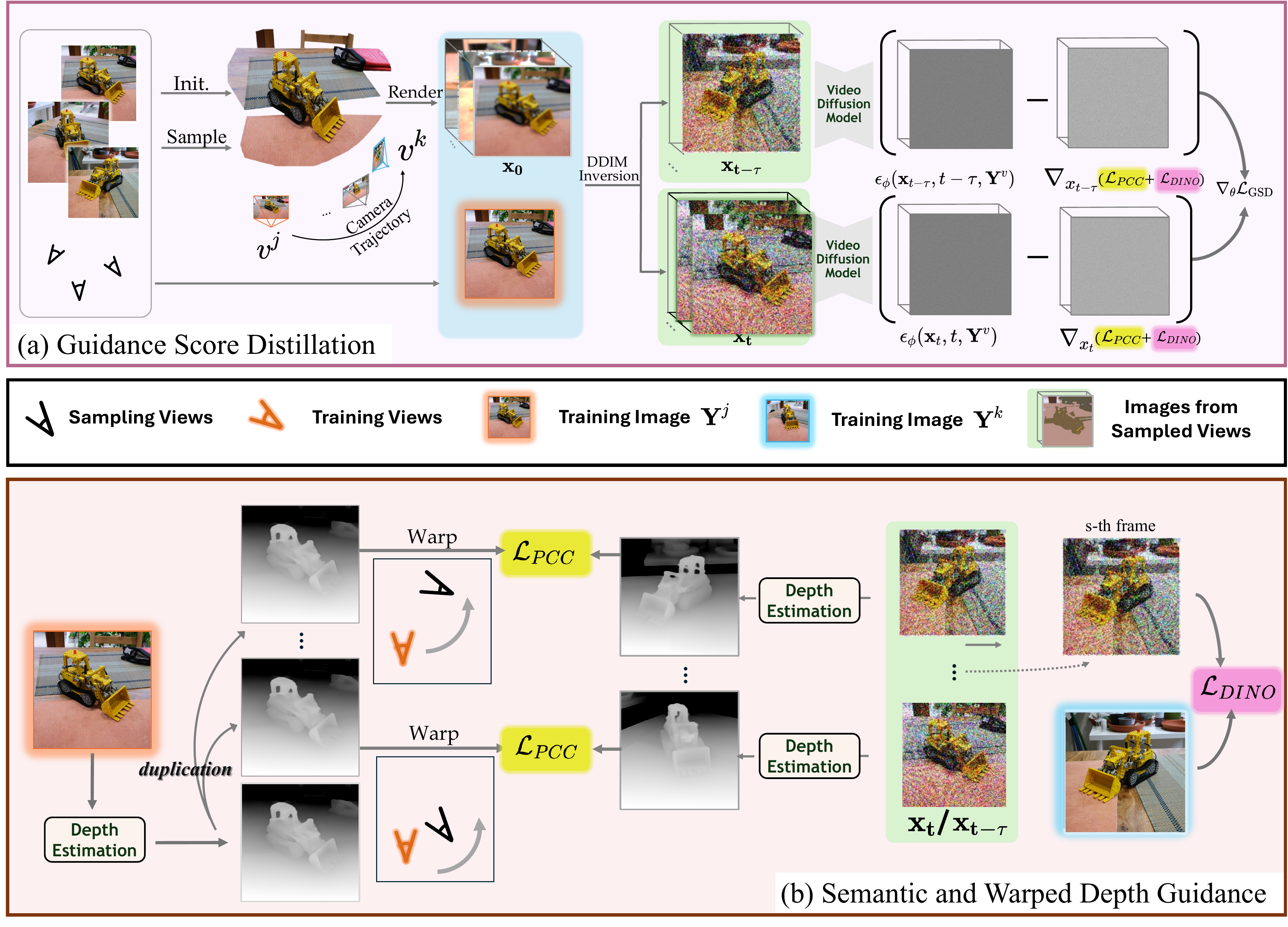}
\caption{The overall framework of our method. Our method first interpolates between $v_j$ and $v_k$ to obtain a set of camera trajectories. Using these trajectories, we render $\mathbf{x}_0$ and apply DDIM inversion to obtain the noise images $\mathbf{x}_t$ and $\mathbf{x}_{t-\tau}$ at two time steps $t$ and $t-\tau$. After predicting the noise for both images using the video diffusion model, we apply $\mathcal{L}_{PCC}$ and $\mathcal{L}_{DINO}$ to correct the noise, which refer to Eq.~\ref{eq:depth} and Eq.~\ref{eq:dino}. Finally, the difference in the corrected results (\emph{i.e.}, $\mathcal{L}_\text{GSD}$) is used to supervise the reconstruction process.
}
\vspace{-4mm}
\end{figure*}

\subsection{Few-shot Novel View Synthesis}
The original NeRF~\cite{mildenhall2021nerf} and 3DGS~\cite{kerbl2023gs} frameworks require hundreds of input views to optimize the 3D representation effectively. However, this is often infeasible in many real-world scenarios. To address these limitations, several pioneering works~\cite{deng2022depth,niemeyer2022regnerf,niemeyer2022regnerf,chung2024depth,jain2021dietnerf,yu2021pixelnerf,zhu2025fsgs} have been proposed to reduce the reliance on the number of training views.
Specifically, most of these approaches add extra constraints to 3D representations. For instance, methods like DepthNeRF~\cite{deng2022depth}, RegNeRF~\cite{niemeyer2022regnerf}, and DNGaussian~\cite{chung2024depth} utilize pretrained monocular depth estimation networks to introduce depth regularization, adding physical constraints to the 3D reconstruction process. Beyond these physical constraints, DietNeRF~\cite{jain2021dietnerf} and pixelNeRF~\cite{yu2021pixelnerf} further enhance training by incorporating regularization in the hidden feature space. More recently, FSGS~\cite{zhu2025fsgs} proposed an innovative strategy to densify Gaussian points during training, which enhances the geometric consistency of the resulting 3D representations. Besides, \cite{xiong2023sparsegs} attempts to introduce 2D diffusion priors to guide the reconstructing process.

% In few-shot novel view synthesis, NeRF-based methods have explored depth regularization~\cite{niemeyer2022regnerf}, frequency annealing~\cite{tancik2020fourier}, and viewpoint distortion~\cite{barron2022mip} to improve performance. These techniques aim to reduce overfitting by regularizing the model and enhancing the consistency of scene reconstructions. However, they still face challenges with extreme data sparsity, often leading to artifacts like blurry or overly smooth textures in the synthesized views.

% For 3DGS, methods like Gaussian densification~\cite{kerbl2023gs}, depth supervision~\cite{wang2023sparsenerf}, and floaters removal in SparseGS~\cite{zhu2023fsgs} also aim to improve the quality of few-shot reconstructions. These approaches help preserve geometric consistency and prevent artifacts, but the lack of intermediate views often results in misalignments and non-representative Gaussians, particularly in complex geometries.

% \zhihao{todo introduce the diffusion based methods}

Despite these advancements, NeRF and 3DGS still struggle with sparse-view inputs, particularly in generating intermediate views with consistent geometry and semantics, leading to suboptimal reconstructions. This highlights the need for more robust approaches to generate new views effectively.

%-------------------------------------------------------------------------
\subsection{Score Distillation}
Score Distillation Sampling (SDS)~\cite{poole2022dreamfusion, wang2023score} extends pre-trained text-to-image diffusion models for sparse-view 3D reconstruction. SDS optimizes an image generator by matching noisy rendered images to distributions learned by a pre-trained diffusion model, improving view consistency in sparse inputs.
While SDS has shown promise, it suffers from issues like oversaturation and a lack of fine details, especially when a high Classifier-Free Guidance (CFG) value is used ~\cite{wang2024prolificdreamer}. To address this, techniques such as time annealing~\cite{lin2023magic3d} have been proposed, gradually reducing the diffusion timesteps to improve the generation quality. Additionally, methods like VSD~\cite{wang2024prolificdreamer} and HiFA~\cite{zhu2023hifa} have reworked the distillation process to better handle noisy inputs and improve image fidelity. Futhermore, \cite{chung2023luciddreamer,lukoianov2024score} propose to apply DDIM inverion to SDS process, enhancing the genenration details. Despite these advancements, SDS-based approaches still struggle with artifacts such as geometry misalignment and overly smooth textures, making it challenging to apply effectively in sparse-view 3D reconstruction tasks. Further refinement is needed for consistent geometry and high realism in low-data scenarios.
\section{Methodology}
\subsection{Preliminary}
\noindent\textbf{Gaussian Splatting.} 3D Gaussian Splatting (3DGS)~\cite{kerbl2023gs} represents points in a 3D scene explicitly through a set of 3D Gaussian points. Each Gaussian point has attributes, including a position vector $\mu \in \mathbb{R}^3$ and a covariance matrix $\Sigma \in \mathbb{R}^{3 \times 3}$. Any point $p$ in 3D space is influenced by these Gaussians, which can be represented by the following 3D Gaussian distribution:
\begin{equation}
    G(p) = \frac{1}{(2\pi)^{3/2} |\Sigma|^{1/2}} e^{-\frac{1}{2} (p - \mu)^T \Sigma^{-1} (p - \mu)}.
\end{equation}
To ensure that the covariance matrix $\Sigma$ is positive semi-definite and physically meaningful, it is typically decomposed into a learnable rotation matrix $R$ and a scaling matrix $S$, such that $\Sigma = R S S^T R^T$. Besides, each Gaussian also stores an opacity logit $o \in \mathbb{R}$ and an appearance feature represented by $n$ spherical harmonic (SH) coefficients $\{\mathbf{c}_i \in \mathbb{R}^3 | i = 1, 2, \dots, n\}$, where $n = D^2$ is the number of SH coefficients, with degree $D$. When rendering, the 2D pixels are obtained by accumulating and weighting the contributions of different Gaussians along the ray direction. Thus, in the rendering pipeline, the complete Gaussian splatting representation $g$ transforms all the learnable parameters $\theta$ to produce the image $x = g(\theta)$. 

The completely differentiable nature of $g(\theta)$ has motivated the use of Differentiable Image Parameterization (DIP)~\cite{mordvintsev2018differentiable} to directly optimize images $x$ rendered from 3D representations. For instance, prior work~\cite{wang2023score,poole2022dreamfusion} has explored constructing loss functions with 2D diffusion models as priors, with one of the most representative groups of methods being Score Distillation~\cite{poole2022dreamfusion,zhu2023hifa,chung2023luciddreamer}.

% The pruning and densification operations~\cite{kerbl2023gs} are also introduced to correct and control the number of Gaussians. However, under sparse view settings, due to insufficient geometric information, these operations tend to optimize the Gaussians only around training views, which often leads to overfitting.

% \zhihao{$\theta$ drive from?}

% Notably, the Gaussian representation in 3DGS is fully differentiable, which inspires the use of Differentiable Image Reparameterization (DIP)~\cite{mordvintsev2018differentiable} technique to directly optimize the images rendered from the 3D representation, where the Gaussian splatting representation $g$ transforms the parameters $\theta$ to produce the image $x = g(\theta)$. Here, $\theta$ denotes the parameter set for the 3D representation, such as the position, size, and orientation of the Gaussian points, which define the structure and properties of the entire 3D scene. For a continuous camera trajectory $\mathbf{p} = \{ p_1, p_2, \dots, p_n \}$ that contains $n$ poses, the rendering process of the corresponding image sequence (or frame sequence) $\mathbf{x}_0$ can be represented as

% To learn the parameters $\theta$, existing work~\cite{wang2023score,poole2022dreamfusion} has explored the use of 2D diffusion models to construct loss functions a prior. One of the most representative method is the Score Distillation~\cite{poole2022dreamfusion,zhu2023hifa,chung2023luciddreamer}.

\noindent\textbf{Score Distillation.} Score distillation is a group of methods that leverages diffusion model priors~\cite{song2020score,ho2020denoising,ho2022classifier,ho2021diffusion} and has achieved significant success across various domains~\cite{song2020score,ho2020denoising,ho2022classifier,ho2021diffusion}. The earliest work in this area is Score Distillation Sampling (SDS)~\cite{poole2022dreamfusion}. Intuitively, SDS adds noise $\epsilon$ to the rendered image $x$, and then uses a pre-trained diffusion model to denoise it. The difference between the predicted noise $\bv{\epsilon_\phi}$ and the real noise $\dg{\epsilon}$ serves as the update direction for the 3D representation, which can be formalized as follows:
\begin{equation}
\nabla_\theta \mathcal{L}_{\text{SDS}}(\theta) = 
\mathbb{E}_{t, \epsilon, c} \left[ \omega(t) (\bv{\epsilon_\phi(x_t, t, y)} - \dg{\epsilon}) \frac{\partial g(\theta, c)}{\partial \theta} \right],
\end{equation}
where the \bv{violet item} and \dg{dark green item} indicate the end and start points of the optimization direction (the same applies hereinafter), respectively.
% This framework can be easily applied to Video Diffusion Models (VDM). Specifically, we aim to use image-to-video models to conditionally generate a video based on the first frame of the input video. A straightforward way to apply SDS to VDM is to select a known view $\mathbf{p}$ and use the corresponding ground truth image $y^{\mathbf{p}}$ as the conditional input in the VDM. Thus, the gradient of the SDS loss can be rewritten as:

% \begin{align}
% \nabla_\theta &\mathcal{L}_{SDS}(\theta)=  \nonumber \\
% &\mathbb{E}_{t, \epsilon, \mathbf{p}} \left[ \omega(t) \underbrace{(\epsilon_\phi(x_t, t, y^{\mathbf{p}}) - \epsilon)}_{\text{SDS Optimization Direction} } \frac{\partial g(\theta, \mathbf{p})}{\partial \theta} \right].
% \end{align}
% unlike generative tasks that encourage diversity, reconstruction or novel view synthesis tasks demand very high geometric precision and color fidelity. In such task settings, some inherent issues with SDS become difficult to ignore.
However, due to the randomness of the noise step $t$, the use of high CFG values, and the generation diversity in the denoising process, the gradients obtained through SDS are often noisy and unstable, which leads to generated results that typically suffer from oversaturation and lack fine details~\cite{poole2022dreamfusion,wang2024prolificdreamer,katzir2023noise}. Fortunately, recent research~\cite{chung2023luciddreamer,lukoianov2024score} proposes modifications to SDS that calculate the difference in predicted noise for images with different noise steps as the optimization direction, where the noise process follows DDIM inversion to produce a deterministic trajectory. This method can effectively mitigate instability issues in SDS. The approach can be summarized as follows:
\begin{align}
&\nabla_{\theta}  \mathcal{L}_{\text{SDS-DDIM}}(\theta) = \nonumber \\ 
&\mathbb{E}_{t, c} \left[ \omega(t)  \left(\bv{\epsilon_{\phi}(x_t, t, y)} - \dg{\epsilon_{\phi}(x_{t - \tau}, t - \tau, y)}\right)  
\frac{\partial g(\theta, c)}{\partial \theta} \right].
\label{eq:sds_ddim}
\end{align}
The $x_t$ is acquired from DDIM inversion~\cite{song2020denoising}
\begin{equation}
 x_{t}=\sqrt{\bar{\alpha}_t}(\hat{x}_0^{t-1}+\frac{\sqrt{1-\bar{\alpha}_t}}{\sqrt{\bar{\alpha}_t}}\boldsymbol{\epsilon}_\phi(x_{t-1},{t-1},y)),
 \label{ddim1}
\end{equation}
where
\begin{equation}
    \hat{x}_0^{t-1}=\frac{x_{t-1}}{\sqrt{\bar{\alpha}_{t-1}}}-\frac{\sqrt{1-\bar{\alpha}_{t-1}}\epsilon_\phi(x_{t-1},{t-1},y)}{\sqrt{\bar{\alpha}_{t-1}}}.
     \label{ddim2}
\end{equation}
By iteratively applying Eq.~\ref{ddim1} to the rendered image $x_0$, we can obtain images $x_t$ and $x_{t-\tau}$ at time steps $t$ and $t-\tau$, respectively. Since Eq.~\ref{ddim1} and Eq.~\ref{ddim2} does not involve the random noise addition, the diffusion trajectory of the images across the two time steps is deterministic and consistent~\cite{chung2023luciddreamer,song2020denoising,lukoianov2024score}. The above framework can be flexibly adapted to any diffusion model.

\subsection{Guidance Score Distillation}
Inspired by the recently popularized Video Diffusion Model (VDM)~\cite{ho2022imagen,guo2023animatediff,blattmann2023stable}, we aim to apply Eq.~\ref{eq:sds_ddim} to VDM to further enhance the multi-view consistency. Specifically, for a continuous camera trajectory $\mathbf{p} = \{ p_1, p_2, \dots, p_n \}$ that contains $n$ poses, we aim to render a sequence of continuous video frames $\mathbf{x}_0 = g(\theta, \mathbf{p})$ as input for the VDM model. At the same time, we ensure that the first frame’s view $p_1$ is set to any one of the training views $v$, and we use the corresponding ground truth image $\mathbf{y}^{v}$ as the condition for the video generation diffusion model. In this case, Eq.~\ref{eq:sds_ddim} can be rewritten as:
\begin{align}
&\nabla_{\theta}  \mathcal{L}_{\text{SDS-DDIM}}^{Video}(\theta) = \nonumber \\ 
&\mathbb{E}_{t, c} \left[ \omega(t)  \left(\bv{\epsilon_{\phi}(\mathbf{x}_t, t, \mathbf{y}^{v})} - \dg{\epsilon_{\phi}(\mathbf{x}_{t - \tau}, t - \tau, \mathbf{y}^{v})}\right)  
\frac{\partial g(\theta, \mathbf{c})}{\partial \theta} \right].
\label{eq:ddim_video}
\end{align}
However, experiments have shown that the above method fail in few-shot novel view synthesis task due to its extremely high requirements for geometric precision and color fidelity. Compared to 2D diffusion models, VDMs introduce new sources of bias, including \emph{object motion within the scene} and \emph{random camera trajectories}. In novel view synthesis tasks, the scene is typically static, and the optimization direction for rendered views needs to strictly align with the camera poses. However, it is unrealistic to expect a VDM to estimate poses from noisy rendered images obtained from  training 3DGS without providing true camera poses explicitly.

% One solution is to use an end-to-end strategy, fine-tuning a pre-trained VDM by real scene camera trajectory videos, enabling the VDM to directly generate static multi-view images from known poses~\cite{reconx,viewcreafter}. However, the high training cost and the fact that these methods only work for specific datasets prevent their widespread use.

To address these issues, instead of fine-tuning the whole pre-trained VDM, which usually costs high, we propose a training-free method for video diffusion models called \emph{Guidance Score Distillation} (GSD). Specifically, we believe that by providing additional guidance $c$ for multiple frames, we can correct the noise predicted by the VDM model $\epsilon_\phi$, thereby reducing or eliminating noise bias, which can be achieved by replacing the original noise $\epsilon_\phi$ with a correction function $\mathbf{F}_t(c, \epsilon_\phi)$ that takes $c$ and $\epsilon_\phi$ as inputs. Thus, Eq.~\ref{eq:ddim_video} can be rewritten in the following new form:
\begin{align}
&\nabla_{\theta}  \mathcal{L}_{\text{GSD}}(\theta) = \nonumber \\ 
&\mathbb{E}_{t, c} \left[ \omega(t)  \left(\bv{\mathbf{F}_t(c, \epsilon_\phi)} - \dg{\mathbf{F}_{t-\tau}(c, \epsilon_\phi)}\right)  
\frac{\partial g(\theta, c)}{\partial \theta} \right].
\label{eq:gsd}
\end{align}
Inspired by~\cite{ho2022classifier,yu2023freedom,zhang2023universal}, the distance measuring function $\mathbf{F}_{t}(c, \epsilon_\phi)$ can be formulated as
\begin{equation}
   \mathbf{F}_{t}(c, \epsilon_\phi) = \epsilon_{\phi}(\mathbf{x}_t, t, \mathbf{y}^{v}) - \gamma(t)\rho_t \nabla_{\mathbf{x}_t} \mathcal{D}_\theta(c, \mathbf{x}_t),
   \label{eq:guidance}
\end{equation}
where $\rho_t$ is a time-dependent coefficient that controls the influence of guidance at different time steps; $\mathcal{D}_\theta (c, \mathbf{x}_t)$ denotes a distance measuring function between the condition $\mathbf{c}$ and the estimated clean image $\mathbf{x}_t$; $\gamma(t)=\frac{\sqrt{\bar{\alpha}_t}}{\sqrt{1-\bar{\alpha}_t}}$ is the conversion factor used to convert image space guidance to noise space guidance. For simplicity, the following equations will use $\lambda_t = \gamma(t)\rho_t$ as the coefficient of $\nabla_{\mathbf{x}_t} \mathcal{D}_\theta(c, \mathbf{x}_t)$. Intuitively, Eq.~\ref{eq:ddim_video} uses the difference between two biased noise predictions at different time steps as the model update direction, while Eq.~\ref{eq:gsd} refines $\epsilon_\phi$ to guide the optimization process with a more accurate noise direction difference. 

Furthermore, we aim to use multiple guidance signals to supervise multiple frames. For the set of guidance $\mathbf{C}=\{\mathbf{c}_1, \mathbf{c}_2, ...,\mathbf{c}_m\}$, we simply assume independence between different guidance, allowing Eq.~\ref{eq:guidance} to contain the sum of multiple distance measurement functions:
\begin{equation}
   \mathbf{F}_{t}(c, \epsilon_\phi) = \epsilon_{\phi}(\mathbf{x}_t, t, \mathbf{y}^{v}) - \lambda_t \nabla_{\mathbf{x}_t} \sum_{i=1}^{m} \eta_i \mathcal{D}_{\theta_i}(c_i, \mathbf{x}_t),
   \label{eq:multi_g}
\end{equation}
where $\eta_i$ denotes the weight of each conditional term. In the following two sections, we will introduce the specific form of $\mathcal{D}_{\theta_i}$ we choose.

\subsection{Semantic Guidance in Distinct Views}
% In this section, we will present two experimentally validated guidance. It should be noted that the guides we provide are tailored to the task settings and challenges of sparse Gaussian Splatting. However, depending on the form of the guidance, the GSD framework could potentially be applied to other tasks, such as 3D editing and 3D stylization.

\begin{figure}[t]
    \scriptsize
    \setlength{\tabcolsep}{0.7pt}
    \centering
    \includegraphics[width=0.42\textwidth]{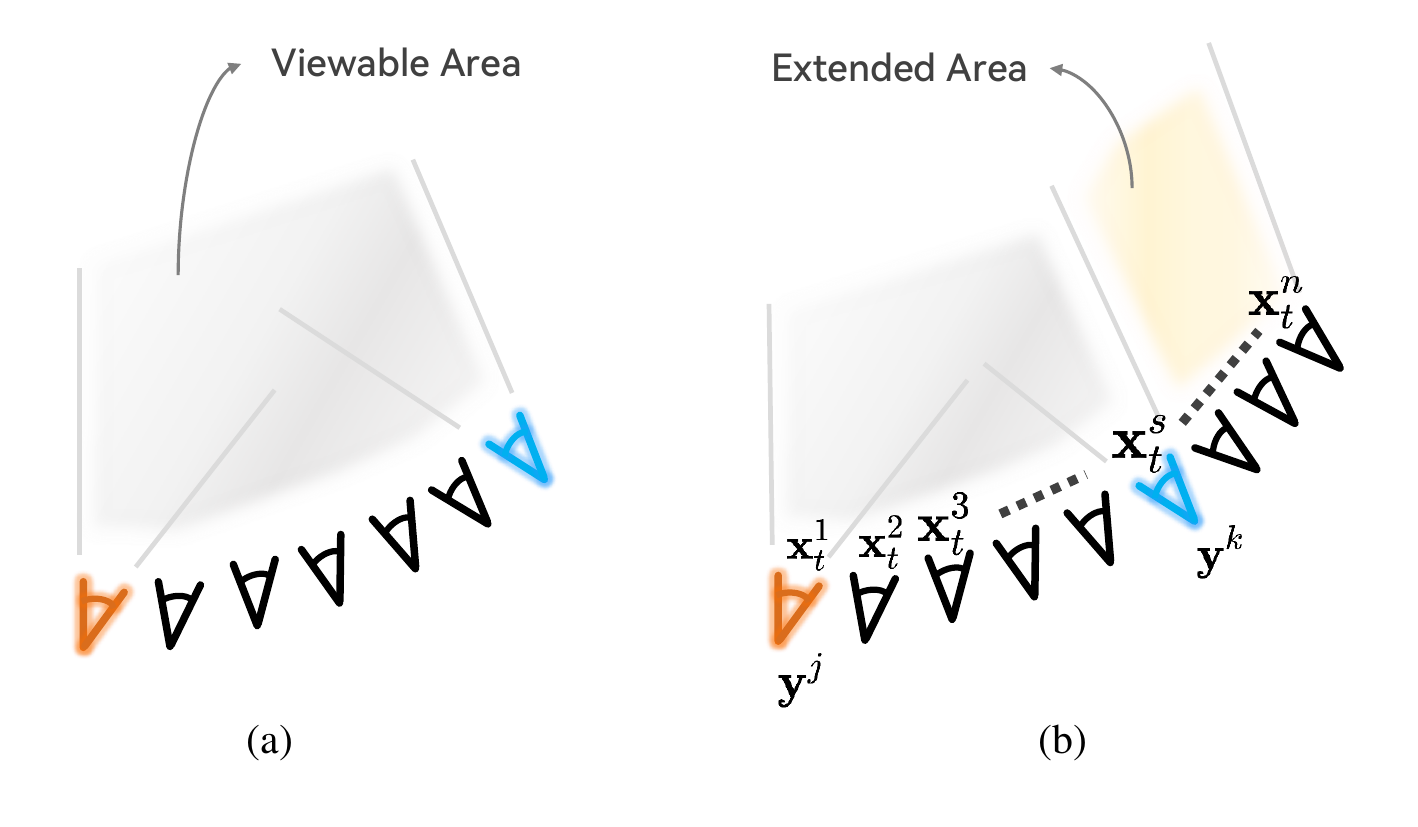}
    \caption{Comparison of two different trajectory generation methods, where (b) obtains a larger viewing area. We also provide the corresponding relation between noisy image $\mathbf{x}_t$ and two ground truth images $\mathbf{y}^j$ and $\mathbf{y}^k$.} 
    \label{view_area}
\end{figure}

% \noindent\textbf{Semantic Feature Guidance.} 
Based on the GSD framework, a naive idea is to add more training camera views in the camera trajectory, thereby guiding the noise direction using multi-frame ground truth images. However, even the best current VDMs can only generate very short videos and cannot adapt to complex camera trajectories. Therefore, we ultimately choose to randomly sample two training camera views $v_j$ and $v_k$ from the training view list $\mathcal{V} = \{v_1, v_2,...,v_r\}$, where $r$ is the number of training views, $v_j$ and $v_k$ (with the corresponding ground truth images $\mathbf{y}^j$ and $\mathbf{y}^k$) serve as the first and last frame in the camera trajectory, respectively. However, in our experiments, we find that this setting may limit the viewable area of the VDM, thus we set $v_k$ as the $s$-th frame of the camera trajectory. Fig.~\ref{view_area} shows their difference. Accordingly, the real image $\mathbf{y}^j$ is used as the condition for the VDM, and $\mathbf{y}^k$ is used as the guidance for correcting the $s$-th frame $\mathbf{x}_t^s$. The distance measuring function of time $t$ for all the training views can be easily expressed as:
\begin{equation}
  \mathcal{D}_{\theta}(\mathbf{y}, \mathbf{x}_t^s) = \mathbb{E}_{(v_j, v_k)\in \mathcal{V} , j\neq k}\|\mathbf{y}^k, \mathbf{x}_t^s\|_1,
\end{equation}
where $\mathcal{V}$ is a set of view combinations, and $\mathbf{Y}$ denotes the training image list. However, our experiments show that directly using the pixel-level loss is not effective. This may be because $\mathcal{L}_1$ equally focuses on the entire image, whereas what most affects the VDM optimization direction should be the semantic features or precise geometric information of the image. Building on the above insights, we use the distance between DINO features~\cite{caron2021emerging} as the distance function, thus the semantic feature guidance can be written:
\begin{equation}
  \mathcal{L}_{GSD}=\mathcal{D}_{\theta}(\mathbf{y}, \mathbf{x}_t^s) = \mathbb{E}_{(v_j, v_k)\in \mathcal{V} , j\neq k}\|M(\mathbf{y}^k), M(\mathbf{x}_t^s)\|_1,
  \label{eq:dino}
\end{equation}
where $M(\cdot)$ refers to the extraction of DINO feature. The semantic features produced by the DINO network effectively utilizes image information, especially helpful in handling long-distance views and complex scenes.

\begin{table*}[t]
    \centering
    \caption{Quantitative results on LLFF~\cite{mildenhall2019llff} with 3, 6, 9 training views. The best, second-best, and third-best entries are marked in red, orange, and yellow, respectively.}
    \vspace{-2mm}
    \begin{tabular}{lccccccccc}
        \hline
        \toprule
        \multirow{2}{*}{\textbf{Method}} & \multicolumn{3}{c}{\textbf{PSNR↑}} & \multicolumn{3}{c}{\textbf{SSIM↑}} & \multicolumn{3}{c}{\textbf{LPIPS↓}} \\
         & 3-view & 6-view & 9-view & 3-view & 6-view & 9-view & 3-view & 6-view & 9-view \\
        \midrule
        Mip-NeRF~\cite{barron2021mipnerf} & 16.11 & 22.91 & 24.88 & 0.401 & 0.756 & 0.826 & 0.460 & 0.213 & 0.160 \\
        DietNeRF~\cite{jain2021dietnerf} & 14.94 & 21.75 & 24.28 & 0.370 & 0.717 & 0.801 & 0.496 & 0.248 & 0.183 \\
        RegNeRF~\cite{niemeyer2022regnerf} & 19.08 & 23.10 & 24.76 & 0.587 & 0.760 & 0.819 & 0.336 & 0.206 & 0.182 \\
        FreeNeRF~\cite{yang2023freenerf} & 19.63 & 23.73 & 25.13 & 0.612 & 0.779 & 0.827 & 0.308 & 0.195 & 0.160 \\
        SparseNeRF~\cite{wang2023sparsenerf} & \cellcolor{yellow!25}19.86 & \cellcolor{yellow!25}23.80 & - & 0.624 & \cellcolor{yellow!25}0.814 & - & 0.328 & \cellcolor{yellow!25}0.125 & - \\
        3DGS~\cite{kerbl2023gs} & 19.22 & \cellcolor{yellow!25}23.80 & \cellcolor{orange!25}25.44 & \cellcolor{yellow!25}0.649 & \cellcolor{yellow!25}0.814 & \cellcolor{yellow!25}0.860 & \cellcolor{yellow!25}0.229 & \cellcolor{yellow!25}0.125 & \cellcolor{orange!25}0.096 \\
        FSGS \cite{zhu2025fsgs} & \cellcolor{orange!25}20.31 &  \cellcolor{orange!25}24.09 & \cellcolor{yellow!25}25.31 & \cellcolor{orange!25}0.652 & \cellcolor{yellow!25}0.823 & \cellcolor{orange!25}0.860 & \cellcolor{yellow!25}0.288 & \cellcolor{orange!25}0.145 & \cellcolor{yellow!25}0.122 \\
        Ours & \cellcolor{red!25}20.58 & \cellcolor{red!25}24.55 & \cellcolor{red!25}25.79 & \cellcolor{red!25}0.673 & \cellcolor{red!25}0.828 & \cellcolor{red!25}0.866 & \cellcolor{red!25}0.208 & \cellcolor{red!25}0.117 & \cellcolor{red!25}0.083 \\
        \bottomrule
        \hline
        \vspace{-3mm}
    \end{tabular}
    \label{llff}
\end{table*}

\subsection{Warped Depth Guidance}
Recent works~\cite{xiong2023sparsegs,xu2022sinnerf} have developed image reprojection techniques that generate pseudo-view images by remapping pixel values in space between different views. 
However, due to the overlap and occlusion between objects, the pseudo-view images generated by this method are often incomplete and lack full semantic information. 
To address this issue, we propose a method that only warps the depth map in space. For a randomly training view $v_j$, we can obtain the relative depth map $D(\mathbf{y}^j)$ via the corresponding ground truth image $\mathbf{y}^j$ as follows,
\begin{equation}
    D(\mathbf{y}^j) = F_m (\mathbf{y}^j),
\end{equation}
where $F_m(\cdot)$ denotes a pretrained monocular depth estimator. For another random view $p^i$ in the camera trajectory, we can similarly estimate the depth map $D(\mathbf{x}_t^i)$ of the rendered image $\mathbf{x}_t^i$. We employ a strategy similar to image reprojection to warp the relative depth $D(y^j)$ from the view $v_j$ to the view $p^i$. It is worth mentioning that we follow the existing method~\cite{xiong2023sparsegs} to estimate the depth range to scale the depth map via the depth of Gaussians. Specifically, for the depth value $d_{\mathbf{y}^j}$ at position $(x, y)$ in $D(\mathbf{y}^j)$, the corresponding depth $d_{\mathbf{x}_t^i}$ can be obtained using the following formula:

\begin{equation}
   d_{\mathbf{x}_t^i} = \text{proj} ( R\cdot ( d_{\mathbf{Y}^j} \cdot K^{-1} \cdot \begin{bmatrix} x \\ y \\ 1 \end{bmatrix} ) + T )_z ,
\end{equation}
where $R$ and $T$ are the rotation matrix and the translation vector between two views; $K^{-1}$ refers to the inverse of the camera intrinsic matrix and $\text{proj}(\cdot)_z$ is the function to project a 3D point back to the 2D image plane and extract the $z$-coordinate from the 3D point after transformation, which corresponds to the depth value. By iterating over all pixels in $\mathbf{y}^j$, we can obtain the warped depth map $\widetilde{D}(\mathbf{y}^j)$ at viewpoint $p^i$, where the areas that are not mapped automatically forms as a $mask(\cdot)$. Thus, we employ a regular depth constraint loss, known as Pearson Correlation Coefficient (PCC)~\cite{xiong2023sparsegs,zhu2025fsgs} loss for the warped depth maps. Applying Eq.~\ref{eq:multi_g}, the distance measuring function can be written as:

\begin{align}
&\mathcal{L}_{depth} = \mathcal{D}_{\theta}(\mathbf{y}, \mathbf{x}_t) = 
\mathbb{E}_{(v_j, v_k)\in \mathcal{V} , j\neq k} \nonumber \\
&\sum_{i=1}^n \left(1 - PCC\left(mask(\widetilde{D}(\mathbf{y}^j)), mask(D(\mathbf{x}_t^i))\right)\right).
  \label{eq:depth}
\end{align}
With warped depth guidance, the prediction noise can be efficiently corrected for each video frame, which provides basic geometric information constraints for the model update direction.

 \begin{figure*}[!t]
		\centering
		\includegraphics[width=14cm, clip, trim=0 0 0 530]{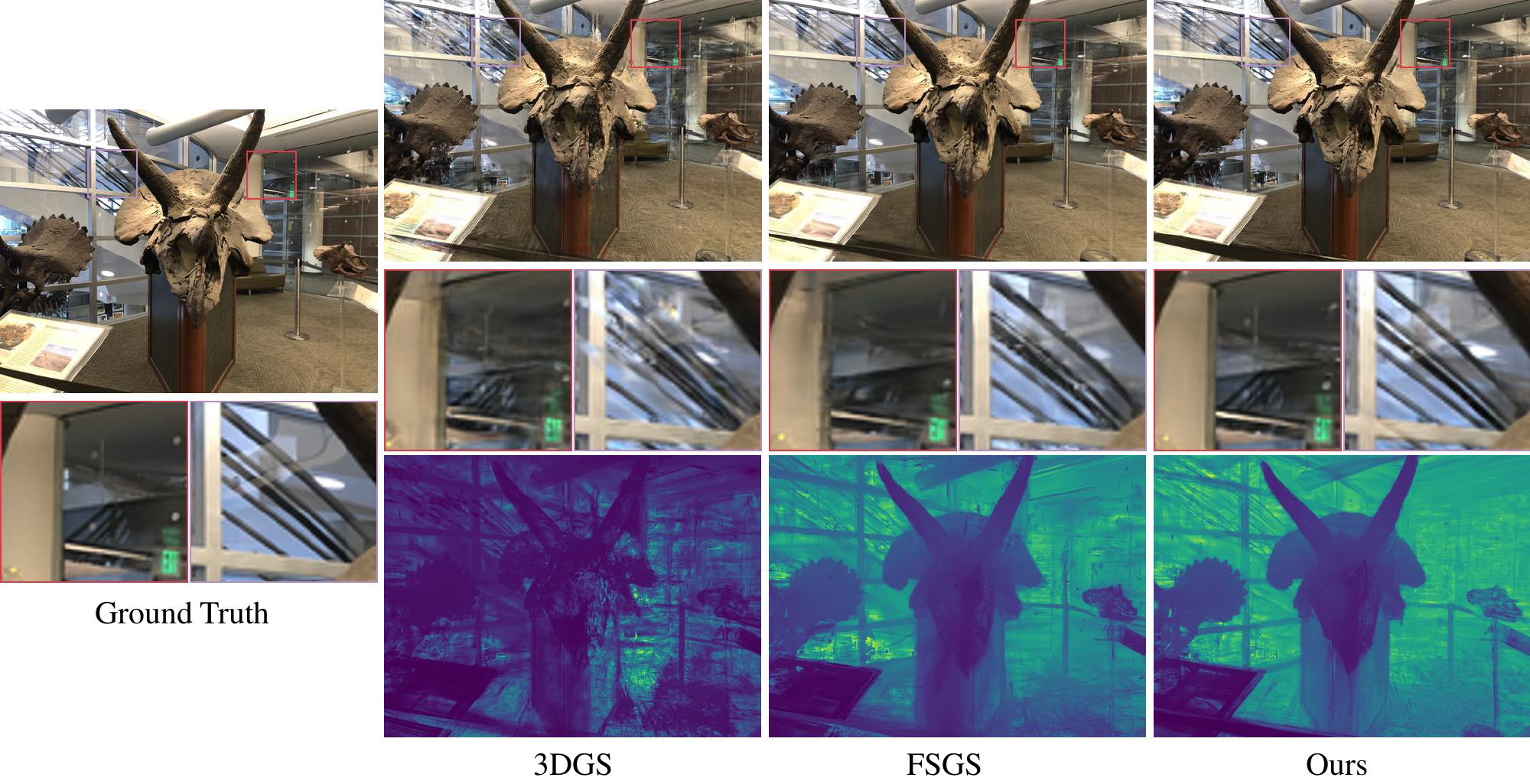}
		\caption{Qualitative comparison on LLFF datasets with 3DGS~\cite{kerbl2023gs}, FSGS~\cite{zhu2025fsgs} and the proposed method. We provide both rendered RGB images and depth maps in the same view for comparsion.}
        \vspace{-1mm}
		\label{llff_contrast}
\end{figure*}

\subsection{Loss Function}
The final total loss formula can be written as
\begin{equation}
    \mathcal{L} = \mathcal{L}_{RGB} + \lambda_{depth}\mathcal{L}_{depth} + \lambda_{GSD} \mathcal{L}_{GSD},
\end{equation}
where $\lambda_{GSD}$ and $\mathcal{L}_{depth}$ are defined in Eq.~\ref{eq:dino} and Eq.~\ref{eq:depth} respectively. In addition, we only use the $\mathcal{L}_{\text{GSD}}$ in the later stages of the 3DGS optimization process, \textit{i.e.}, after 3000 iterations, as depth range estimation quality typically improves after the initial optimization phase.

 \begin{figure*}[!t]
		\centering
		\includegraphics[width=14cm, height=7.2cm]{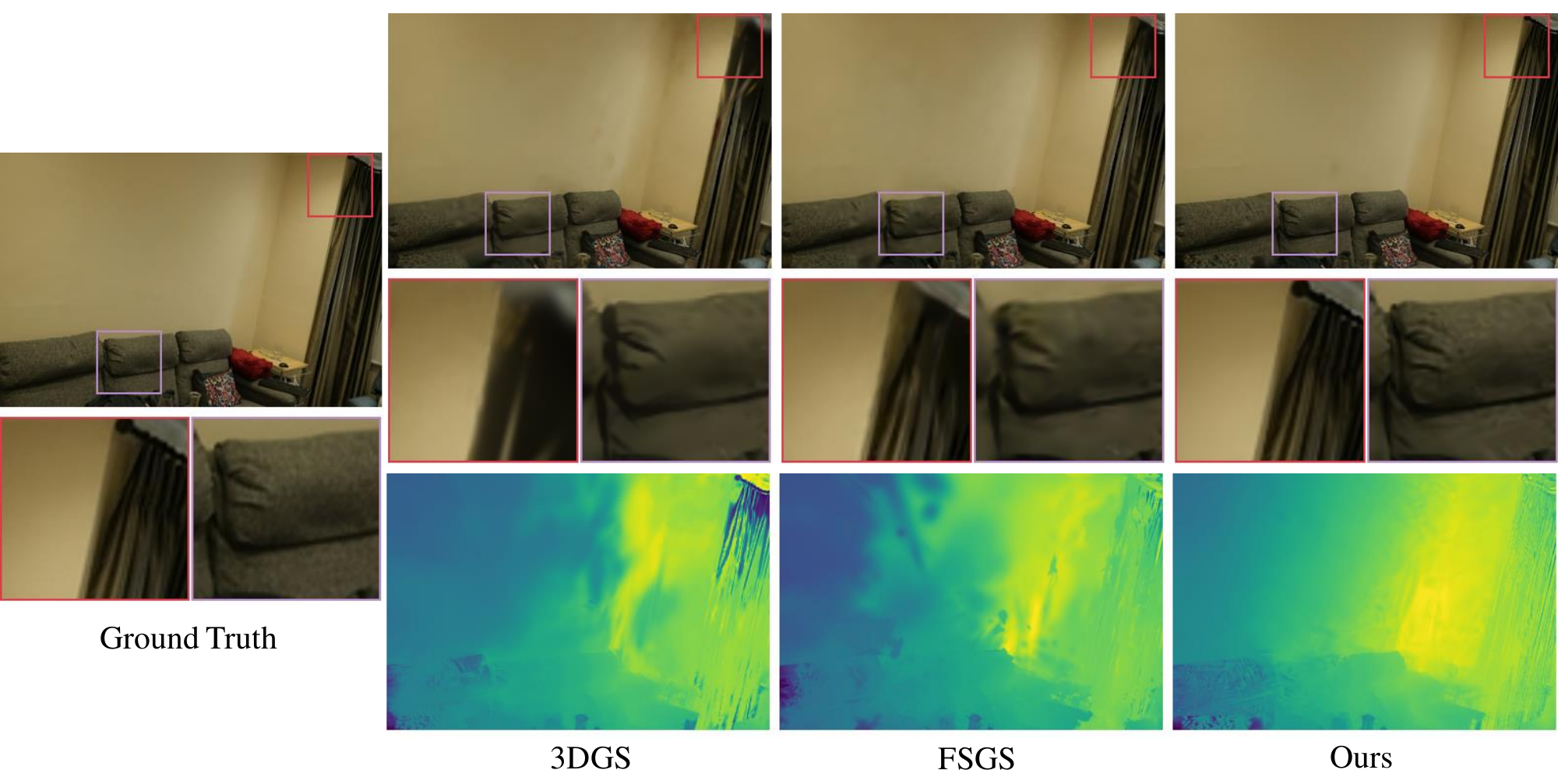}
		\caption{Qualitative comparison on MipNeRF360 datasets with 3DGS~\cite{kerbl2023gs}, FSGS~\cite{zhu2025fsgs} and the proposed method. We provide both rendered RGB images and depth maps in the same view for comparsion.}
		\label{mipnerf360_contrast}
\end{figure*}

\section{Experiment}

\subsection{Setup}

\noindent\textbf{Datasets} We conducted experiments on three datasets: LLFF~\cite{mildenhall2019llff}, Mip-NeRF360~\cite{barron2022mipnerf360} and DTU~\cite{jensen2014dtu}. For LLFF datasets~\cite{mildenhall2019llff}, we select every eighth image as the test set, and evenly sample sparse views from the remaining images for training. Besides, we use an 8x downsampling rate to train on 3, 6, and 9 input views, respectively. For the Mip-NeRF360 dataset, we train on 24 input views with 8x and 4x downsampling rates, respectively. For the DTU dataset, an 8-view configuration is trained with a 4x downsampling rate. To focus on the target object and reduce background noise during evaluation, we apply object masks to DTU, similar to prior works~\cite{niemeyer2022regnerf}. 

\noindent\textbf{Baselines.} We compare GSD with several few-shot NVS methods on these datasets, including DietNeRF~\cite{jain2021dietnerf}, RegNeRF~\cite{niemeyer2022regnerf}, FreeNeRF~\cite{yang2023freenerf}, and SparseNeRF~\cite{wang2023sparsenerf}. Additionally, we include comparisons with the high-performing Mip-NeRF~\cite{barron2021mipnerf}, primarily designed for dense-view training, and point-based 3DGS~\cite{kerbl2023gs}, following its original dense-view training recipe. Following~\cite{xiong2023sparsegs,zhu2025fsgs,niemeyer2022regnerf}, we report the average PSNR, SSIM, LPIPS scores for all the methods.

\noindent\textbf{Settings.} The video diffusion model we adopt is Stable Video Diffusion~\cite{blattmann2023stable}. We implemented GSD by the PyTorch framework, with the initial point cloud computed from SfM only with the training views. During training, we also applied the conventional densification and pruning operations~\cite{zhu2025fsgs,kerbl2023gs}. We densify the Gaussians every 100 iterations and start densification after 500 iterations. The total optimization steps are set to 10,000 on LLFF and DTU datasets and 30,000 on MipNeRF360 datasets. We start to use the $\mathcal{L}_{GSD}$ after 3,000 iterations. We utilize the pre-trained DPT model~\cite{Ranftl2021} for depth estimation. All results are obtained using an NVIDIA A40 GPU.
\begin{table}[ht]
\setlength{\tabcolsep}{0.1mm}
    \small
    \centering
    \caption{Quantitative results on Mip-NeRF360 datasets~\cite{barron2022mipnerf360} at 1/8 and 1/4 resolutions. The best, second-best, and third-best entries are marked in red, orange, and yellow, respectively.}
    \begin{tabular}{lcccccc}
        \hline \toprule
        \multirow{2}{*}{\textbf{Methods}} & \multicolumn{3}{c}{\textbf{1/8 Resolution}} & \multicolumn{3}{c}{\textbf{1/4 Resolution}} \\
         & \textbf{PSNR↑} & \textbf{SSIM↑} & \textbf{LPIPS↓} & \textbf{PSNR↑} & \textbf{SSIM↑} & \textbf{LPIPS↓} \\
        \midrule
        Mip-NeRF360~\cite{barron2022mipnerf360} & 21.23 & 0.613 & 0.351 & 19.78 & 0.530 & 0.431 \\
        
        RegNeRF~\cite{niemeyer2022regnerf} & 22.19 & 0.643 & 0.335 & 20.55 & 0.546 & 0.398 \\
        FreeNeRF~\cite{yang2023freenerf} & 22.78 & 0.689 & \cellcolor{yellow!25}0.323 & \cellcolor{yellow!25}21.04 & 0.587 & 0.377 \\
        SparseNeRF~\cite{wang2023sparsenerf} & \cellcolor{yellow!25}22.85 & \cellcolor{yellow!25}0.693 & 0.315 & 21.13 & \cellcolor{yellow!25}0.600 & 0.389 \\
        
        DietNeRF~\cite{jain2021dietnerf} & 20.21 & 0.557 & 0.387 & 19.11 & 0.482 & 0.452 \\
        3D-GS~\cite{kerbl2023gs} & 20.89 & 0.633 & 0.317 & 19.93 & 0.588 & 0.401 \\
        FSGS~\cite{zhu2025fsgs} & \cellcolor{orange!25}23.70 & \cellcolor{orange!25}0.745 & \cellcolor{orange!25}0.220 & \cellcolor{orange!25}22.82 & \cellcolor{red!25}0.693 & \cellcolor{orange!25}0.293 \\
        Ours & \cellcolor{red!25}23.74 & \cellcolor{red!25}0.755 & \cellcolor{red!25}0.208 & \cellcolor{red!25}22.95 & \cellcolor{orange!25}0.691 & \cellcolor{red!25}0.277 \\
        \bottomrule 
        \hline 
    \end{tabular}
    \label{mipnerf360}
\end{table}

\begin{table}[ht]
    \centering
    \caption{Quantitative results on DTU~\cite{jensen2014dtu} with 3 training views. The best, second-best, and third-best entries are marked in red, orange, and yellow, respectively.}
    \vspace{-3mm}
    \begin{tabular}{lccc}
        \hline
        \toprule
        \textbf{Method} & \textbf{PSNR↑} & \textbf{SSIM↑} & \textbf{LPIPS↓} \\
        \midrule
        DietNeRF~\cite{jain2021dietnerf} & 11.85 & 0.633 & 0.314 \\
        RegNeRF~\cite{niemeyer2022regnerf} & 18.89 & 0.745 & 0.190 \\
        Mip-NeRF~\cite{barron2021mipnerf} & 9.10 & 0.578 & 0.348 \\
        FreeNeRF~\cite{yang2023freenerf} & \cellcolor{red!25}19.92 & \cellcolor{yellow!25}0.787 & \cellcolor{orange!25}0.182 \\
        SparseNeRF~\cite{wang2023sparsenerf} & \cellcolor{orange!25}19.55 & - & 0.201 \\
        3DGS~\cite{kerbl2023gs} & 17.65 & \cellcolor{red!25}0.846 & \cellcolor{yellow!25}0.146 \\
        Ours & \cellcolor{yellow!25}19.33 & \cellcolor{orange!25}0.842 & \cellcolor{red!25}0.117 \\
        \bottomrule
        \hline
    \end{tabular}
    \label{dtu}
\end{table}
\begin{table}[ht]
    \centering
    \caption{Ablation study of the second camera guidance on the fern scene of the LLFF dataset. The best, second-best, and third-best entries are marked in red, orange, and yellow, respectively.}
    \vspace{-3mm}
    \begin{tabular}{lccc}
        \hline
        \toprule
        \textbf{Method} & \textbf{PSNR↑} & \textbf{SSIM↑} & \textbf{LPIPS↓} \\
        \midrule
        w/o 2nd camera guidance & 21.81 & 0.68 & \cellcolor{orange!25}0.24 \\
        pixel level guidance & 21.81 & \cellcolor{yellow!25}0.69 & \cellcolor{orange!25}0.21 \\
        Ours & \cellcolor{red!25}21.99 & \cellcolor{red!25}0.74 & \cellcolor{red!25}0.20 \\
        \bottomrule
        \hline
    \end{tabular}
    \label{cam_guidance}
\end{table}
\subsection{Comparisons}

\noindent\textbf{LLFF datasets.} We present the quantitative results on the LLFF dataset in Tab~\ref{llff}. Our method consistently achieves the best performance across PSNR, SSIM, and LPIPS metrics with 3, 6, and 9 training views. Depth-supervised FSGS~\cite{zhu2025fsgs} enhances the performance of 3DGS with 3 and 6 views, but cannot avoid geometric ambiguities with more views. Our method incorporates view-consistent video diffusion model priors to prevent the reconstruction of incorrect geometry, leading to improvements across all metrics with various training views. Fig.~\ref{llff_contrast} provides quantitative visualizations of rendered images and depth maps at novel views and Gaussian points. From the rendered depth maps, we observe that the original 3DGS exhibits significant errors in reconstructing distant regions. With depth supervision, FSGS can correct the erroneous depth of 3DGS but still struggles with the reconstruction of image details, such as distant leaves and railings. Our method effectively recovered photometric details, resulting in higher-quality outcomes.

\noindent\textbf{MipNeRF360 datasets.} The quantitative results on the Mip-NeRF360 dataset are provided in Tab~\ref{mipnerf360}. Our method consistently achieves the best or second-best performance in PSNR, SSIM, and LPIPS metrics at downsampling rates of 1/8 and 1/4. Notably, the performance improvement brought by our method decreases at higher resolutions, which may be due to the fact that the video diffusion model~\cite{blattmann2023stable} we used is trained at a lower resolution, leading to lower-quality gradient directions at higher resolutions. Fig.~\ref{mipnerf360_contrast} presents quantitative visualizations of rendered images and depth at novel views and Gaussian points. From the figure, we can observe that our method achieves better reconstruction quality at the edges of the scene, which partly benefits from our extended sampling strategy for the camera trajectory.

\noindent\textbf{DTU datasets.} The quantitative results on the DTU dataset are provided in Tab.~\ref{dtu}. Our method achieves competitive results. It is worth mentioning that, due to the use of generative model priors, our method consistently maintains the best performance in LPIPS. The relatively lower results in other metrics may be because the training dataset of the video diffusion model contains very few examples of single objects with solid colored backgrounds, which presents a challenge for its generation capabilities.

\begin{table}[ht]
    \centering
    \caption{Ablation study of the camera trajectory sampling method on the fern scene of the LLFF dataset.}
    \vspace{-3mm}
    \begin{tabular}{lccc}
        \hline
        \toprule
        \textbf{Method} & \textbf{PSNR↑} & \textbf{SSIM↑} & \textbf{LPIPS↓} \\
        \midrule
        limited camera trajectory & 21.94 & 0.70 & 0.24 \\
        Ours & 21.99 & 0.74 & 0.20 \\
        \bottomrule
        \hline
    \end{tabular}
    \label{cam_track}
\end{table}

\begin{table}[ht]
    \centering
    \caption{Ablation study on the fern scene of the LLFF dataset. The best, second-best, and third-best entries are marked in red, orange, and yellow, respectively.}
    \vspace{-3mm}
    \begin{tabular}{lccc}
        \hline
        \toprule
        \textbf{Method} & \textbf{PSNR↑} & \textbf{SSIM↑} & \textbf{LPIPS↓} \\
        \midrule
        w/o 2nd camera guidance & \cellcolor{yellow!25}21.81 & \cellcolor{yellow!25}0.68 & \cellcolor{orange!25}0.24 \\
        w/o warp depth guidance & \cellcolor{orange!25}21.83 & \cellcolor{orange!25}0.71 & \cellcolor{yellow!25}0.29 \\
        w/o all guidance & 21.58 & 0.64 & 0.33 \\
        w/o SDS-DDIM & 21.52 & 0.66 & 0.31 \\
        Ours & \cellcolor{red!25}21.99 & \cellcolor{red!25}0.74 & \cellcolor{red!25}0.20 \\
        \bottomrule
        \hline
    \end{tabular}
    \label{ablation_all}
\end{table}

\subsection{Ablation Study}

\noindent\textbf{The second camera guidance.} We conducted an ablation study on the guidance method for the second camera. We compared semantic-level guidance, pixel-level guidance, and no guidance (see Tab.~\ref{cam_guidance}). The experiments indicate that our method achieves optimal performance. Pixel-level guidance may not effectively help correct the noise bias of the latent diffusion model.

\noindent\textbf{The camera sampling method.} We also performed an ablation study on the sampling method for camera trajectories (see Tab.~\ref{cam_track}), confirming that expansive sampling can broaden the visible range and improve model performance.

\noindent\textbf{Others components of the model.} Furthermore, we conducted an ablation study on the components of the model (see Tab.~\ref{ablation_all}), confirming that both types of guidance we designed play a role. Comparatively, the improvements brought by the guidance for the second camera are more significant. Additional ablation studies and hyperparameter discussions can be found in the supplementary materials.

\section{Conclusion}

In this paper, we propose the Guidance Score Distillation (GSD) framework to address overfitting in 3D Gaussian Splatting (3DGS) when dealing with few-shot views. Our approach leverages pretrained video diffusion models (VDM) to extract multi-view consistency priors, applying score distillation sampling (SDS) to guide the Gaussian representation toward VDM’s generative direction. We incorporate warping guidance based on real depth maps and additional guidance based on semantic image features to ensure alignment in geometry and camera poses. Experimental results demonstrate that GSD effectively enhances the performance of 3DGS and achieves superior results across multiple datasets.
{
    \small
    \bibliographystyle{ieeenat_fullname}
    \bibliography{main}

@String(CVPR= {IEEE Conf. Comput. Vis. Pattern Recog.})

@String(ICCV= {Int. Conf. Comput. Vis.})

@String(TOG= {ACM Trans. Graph.})

@String(ICLR = {Int. Conf. Learn. Represent.})

@String(VR   = {Vis. Res.})

@String(CVPR  = {CVPR})

@String(ICCV  = {ICCV})

@String(TOG   = {ACM TOG})

@String(ICLR  = {ICLR})

@article{kerbl2023gs,
  title={3D Gaussian Splatting for Real-Time Radiance Field Rendering},
  author={Kerbl, Bernhard and Reiser, Christian and Liao, Zexiang and Kopanas, Georgios and Zhang, Thomas and Leimk{\"u}hler, Thomas and Parger, Matthias and Rousselle, Fabrice and Nie{\ss}ner, Matthias and Kellnhofer, Petr},
  journal={ACM Transactions on Graphics (TOG)},
  volume={42},
  number={4},
  pages={1--12},
  year={2023},
  publisher={ACM New York, NY}
}

@article{xiong2023sparsegs,
  title={SparseGS: Real-Time 360° Sparse View Synthesis using Gaussian Splatting},
  author={Xiong, Haolin and Muttukuru, Sairisheek and Upadhyay, Rishi and Chari, Pradyumna and Kadambi, Achuta},
  journal={arXiv preprint arXiv:2312.00206},
  year={2023}
}

@article{mordvintsev2018differentiable,
  author = {Mordvintsev, Alexander and Pezzotti, Nicola and Schubert, Ludwig and Olah, Chris},
  title = {Differentiable Image Parameterizations},
  journal = {Distill},
  year = {2018},
  note = {https://distill.pub/2018/differentiable-parameterizations},
  doi = {10.23915/distill.00012}
}

@inproceedings{wang2023score,
  title={Score jacobian chaining: Lifting pretrained 2d diffusion models for 3d generation},
  author={Wang, Haochen and Du, Xiaodan and Li, Jiahao and Yeh, Raymond A and Shakhnarovich, Greg},
  booktitle={Proceedings of the IEEE/CVF Conference on Computer Vision and Pattern Recognition},
  pages={12619--12629},
  year={2023}
}

@article{poole2022dreamfusion,
  title={Dreamfusion: Text-to-3d using 2d diffusion},
  author={Poole, Ben and Jain, Ajay and Barron, Jonathan T and Mildenhall, Ben},
  journal={arXiv preprint arXiv:2209.14988},
  year={2022}
}

@inproceedings{barron2021mipnerf,
  title={{Mip-NeRF}: A Multiscale Representation for Anti-Aliasing Neural Radiance Fields},
  author={Barron, Jonathan T. and Mildenhall, Ben and Tancik, Matthew and Hedman, Peter and Martin-Brualla, Ricardo and Srinivasan, Pratul P.},
  booktitle={Proceedings of the IEEE/CVF International Conference on Computer Vision (ICCV)},
  pages={5855--5864},
  year={2021}
}

@inproceedings{jain2021dietnerf,
  title={{DietNeRF}: Few-Shot Novel View Synthesis via Diffusion of Viewpoints},
  author={Jain, Ajay and Tancik, Matthew and Abbeel, Pieter},
  booktitle={Proceedings of the IEEE/CVF Conference on Computer Vision and Pattern Recognition (CVPR)},
  pages={7226--7235},
  year={2021}
}

@inproceedings{yang2023freenerf,
  title={{FreeNeRF}: Improving Few-Shot Neural Rendering with Free Frequency Regularization},
  author={Yang, Jingyang and Zhang, Yinda and Li, Zhoutong and Zhang, Qi and Zhang, Jianmin and Bao, Hujun},
  booktitle={Proceedings of the IEEE/CVF Conference on Computer Vision and Pattern Recognition (CVPR)},
  pages={18624--18633},
  year={2023}
}

@inproceedings{wang2023sparsenerf,
  title={{SparseNeRF}: Distilling Depth Ranking for Few-Shot Novel View Synthesis},
  author={Wang, Cong and Zhang, Yinda and Li, Zhoutong and Zhang, Qi and Zhang, Jianmin and Bao, Hujun},
  booktitle={Proceedings of the IEEE/CVF International Conference on Computer Vision (ICCV)},
  pages={1234--1243},
  year={2023}
}

@inproceedings{niemeyer2022regnerf,
  title={{RegNeRF}: Regularizing Neural Radiance Fields for View Synthesis from Sparse Inputs},
  author={Niemeyer, Michael and Barron, Jonathan T. and Mildenhall, Ben and Sajjadi, Mehdi S. M. and Geiger, Andreas and Radwan, Noha},
  booktitle={Proceedings of the IEEE/CVF Conference on Computer Vision and Pattern Recognition (CVPR)},
  pages={5480--5490},
  year={2022}
}

@inproceedings{ho2021diffusion,
  title={Diffusion Models Beat GANs on Image Synthesis},
  author={Jonathan Ho and Ajay Jain and Pieter Abbeel},
  booktitle={NeurIPS 2021},
  year={2021}
}

@inproceedings{zhu2025fsgs,
  title={Fsgs: Real-time few-shot view synthesis using gaussian splatting},
  author={Zhu, Zehao and Fan, Zhiwen and Jiang, Yifan and Wang, Zhangyang},
  booktitle={European Conference on Computer Vision},
  pages={145--163},
  year={2025},
  organization={Springer}
}

@inproceedings{ho2022classifier,
  title={Classifying Diffusion Models with Classifier Guidance},
  author={Jonathan Ho and Tim Salimans and Niru Maheswaranathan and Mohammad Amin Sadeghi and Kevin Murphy},
  booktitle={NeurIPS 2022},
  year={2022}
}

@inproceedings{ho2020denoising,
  title={Denoising Diffusion Probabilistic Models},
  author={Jonathan Ho and Ajay Jain and Pieter Abbeel},
  booktitle={NeurIPS 2020},
  year={2020}
}

@inproceedings{song2020score,
  title={Score-Based Generative Modeling through Stochastic Differential Equations},
  author={Yang Song and Stefano Ermon},
  booktitle={ICLR 2021},
  year={2021}
}

@article{chung2023luciddreamer,
  title={Luciddreamer: Domain-free generation of 3d gaussian splatting scenes},
  author={Chung, Jaeyoung and Lee, Suyoung and Nam, Hyeongjin and Lee, Jaerin and Lee, Kyoung Mu},
  journal={arXiv preprint arXiv:2311.13384},
  year={2023}
}

@article{lukoianov2024score,
  title={Score Distillation via Reparametrized DDIM},
  author={Lukoianov, Artem and Borde, Haitz S{\'a}ez de Oc{\'a}riz and Greenewald, Kristjan and Guizilini, Vitor Campagnolo and Bagautdinov, Timur and Sitzmann, Vincent and Solomon, Justin},
  journal={arXiv preprint arXiv:2405.15891},
  year={2024}
}

@article{katzir2023noise,
  title={Noise-free score distillation},
  author={Katzir, Oren and Patashnik, Or and Cohen-Or, Daniel and Lischinski, Dani},
  journal={arXiv preprint arXiv:2310.17590},
  year={2023}
}

@article{wang2023steindreamer,
  title={Steindreamer: Variance reduction for text-to-3d score distillation via stein identity},
  author={Wang, Peihao and Fan, Zhiwen and Xu, Dejia and Wang, Dilin and Mohan, Sreyas and Iandola, Forrest and Ranjan, Rakesh and Li, Yilei and Liu, Qiang and Wang, Zhangyang and others},
  journal={arXiv preprint arXiv:2401.00604},
  year={2023}
}

@article{zhu2023hifa,
  title={Hifa: High-fidelity text-to-3d generation with advanced diffusion guidance},
  author={Zhu, Junzhe and Zhuang, Peiye and Koyejo, Sanmi},
  journal={arXiv preprint arXiv:2305.18766},
  year={2023}
}

@article{wang2024prolificdreamer,
  title={Prolificdreamer: High-fidelity and diverse text-to-3d generation with variational score distillation},
  author={Wang, Zhengyi and Lu, Cheng and Wang, Yikai and Bao, Fan and Li, Chongxuan and Su, Hang and Zhu, Jun},
  journal={Advances in Neural Information Processing Systems},
  volume={36},
  year={2024}
}

@inproceedings{caron2021emerging,
  title={Emerging properties in self-supervised vision transformers},
  author={Caron, Mathilde and Touvron, Hugo and Misra, Ishan and J{\'e}gou, Herv{\'e} and Mairal, Julien and Bojanowski, Piotr and Joulin, Armand},
  booktitle={Proceedings of the IEEE/CVF international conference on computer vision},
  pages={9650--9660},
  year={2021}
}

@article{mildenhall2021nerf,
  title={Nerf: Representing scenes as neural radiance fields for view synthesis},
  author={Mildenhall, Ben and Srinivasan, Pratul P and Tancik, Matthew and Barron, Jonathan T and Ramamoorthi, Ravi and Ng, Ren},
  journal={Communications of the ACM},
  volume={65},
  number={1},
  pages={99--106},
  year={2021},
  publisher={ACM New York, NY, USA}
}

@article{tancik2020fourier,
  title={Fourier features let networks learn high frequency functions in low dimensional domains},
  author={Tancik, Matthew and Srinivasan, Pratul and Mildenhall, Ben and Fridovich-Keil, Sara and Raghavan, Nithin and Singhal, Utkarsh and Ramamoorthi, Ravi and Barron, Jonathan and Ng, Ren},
  journal={Advances in neural information processing systems},
  volume={33},
  pages={7537--7547},
  year={2020}
}

@inproceedings{barron2022mip,
  title={Mip-nerf 360: Unbounded anti-aliased neural radiance fields},
  author={Barron, Jonathan T and Mildenhall, Ben and Verbin, Dor and Srinivasan, Pratul P and Hedman, Peter},
  booktitle={Proceedings of the IEEE/CVF conference on computer vision and pattern recognition},
  pages={5470--5479},
  year={2022}
}

@article{masoumian2022monocular,
  title={Monocular depth estimation using deep learning: A review},
  author={Masoumian, Armin and Rashwan, Hatem A and Cristiano, Juli{\'a}n and Asif, M Salman and Puig, Domenec},
  journal={Sensors},
  volume={22},
  number={14},
  pages={5353},
  year={2022},
  publisher={MDPI}
}

@inproceedings{zhan2018unsupervised,
  title={Unsupervised learning of monocular depth estimation and visual odometry with deep feature reconstruction},
  author={Zhan, Huangying and Garg, Ravi and Weerasekera, Chamara Saroj and Li, Kejie and Agarwal, Harsh and Reid, Ian},
  booktitle={Proceedings of the IEEE conference on computer vision and pattern recognition},
  pages={340--349},
  year={2018}
}

@inproceedings{lin2023magic3d,
  title={Magic3d: High-resolution text-to-3d content creation},
  author={Lin, Chen-Hsuan and Gao, Jun and Tang, Luming and Takikawa, Towaki and Zeng, Xiaohui and Huang, Xun and Kreis, Karsten and Fidler, Sanja and Liu, Ming-Yu and Lin, Tsung-Yi},
  booktitle={Proceedings of the IEEE/CVF Conference on Computer Vision and Pattern Recognition},
  pages={300--309},
  year={2023}
}

@article{song2020denoising,
  title={Denoising diffusion implicit models},
  author={Song, Jiaming and Meng, Chenlin and Ermon, Stefano},
  journal={arXiv preprint arXiv:2010.02502},
  year={2020}
}

@article{Ranftl2021,
	author    = {Ren\'{e} Ranftl and Alexey Bochkovskiy and Vladlen Koltun},
	title     = {Vision Transformers for Dense Prediction},
	journal   = {ArXiv preprint},
	year      = {2021},
}

@inproceedings{cho2019novel,
  title={Novel view synthesis with multiple 360 images for large-scale 6-dof virtual reality system},
  author={Cho, Hochul and Kim, Jangyoon and Woo, Woontack},
  booktitle={2019 IEEE Conference on Virtual Reality and 3D User Interfaces (VR)},
  pages={880--881},
  year={2019},
  organization={IEEE}
}

@inproceedings{tseng2022pseudo,
author = {Tseng, Kuan-Wei and Huang, Jing-Yuan and Chen, Yang-Shen and Chen, Chu-Song and Hung, Yi-Ping},
title = {Pseudo-3D Scene Modeling for Virtual Reality Using Stylized Novel View Synthesis},
year = {2022},
isbn = {9781450393614},
publisher = {Association for Computing Machinery},
address = {New York, NY, USA},
url = {https://doi.org/10.1145/3532719.3543232},
doi = {10.1145/3532719.3543232},
booktitle = {ACM SIGGRAPH 2022 Posters},
articleno = {66},
numpages = {2},
location = {Vancouver, BC, Canada},
series = {SIGGRAPH '22}
}

@inproceedings{tonderski2024neurad,
  title={Neurad: Neural rendering for autonomous driving},
  author={Tonderski, Adam and Lindstr{\"o}m, Carl and Hess, Georg and Ljungbergh, William and Svensson, Lennart and Petersson, Christoffer},
  booktitle={Proceedings of the IEEE/CVF Conference on Computer Vision and Pattern Recognition},
  pages={14895--14904},
  year={2024}
}

@inproceedings{huang2023neural,
  title={Neural lidar fields for novel view synthesis},
  author={Huang, Shengyu and Gojcic, Zan and Wang, Zian and Williams, Francis and Kasten, Yoni and Fidler, Sanja and Schindler, Konrad and Litany, Or},
  booktitle={Proceedings of the IEEE/CVF International Conference on Computer Vision},
  pages={18236--18246},
  year={2023}
}

@article{watson2022novel,
  title={Novel view synthesis with diffusion models},
  author={Watson, Daniel and Chan, William and Martin-Brualla, Ricardo and Ho, Jonathan and Tagliasacchi, Andrea and Norouzi, Mohammad},
  journal={arXiv preprint arXiv:2210.04628},
  year={2022}
}

@inproceedings{voleti2025sv3d,
  title={Sv3d: Novel multi-view synthesis and 3d generation from a single image using latent video diffusion},
  author={Voleti, Vikram and Yao, Chun-Han and Boss, Mark and Letts, Adam and Pankratz, David and Tochilkin, Dmitry and Laforte, Christian and Rombach, Robin and Jampani, Varun},
  booktitle={European Conference on Computer Vision},
  pages={439--457},
  year={2025},
  organization={Springer}
}

@inproceedings{deng2022depth,
  title={Depth-supervised nerf: Fewer views and faster training for free},
  author={Deng, Kangle and Liu, Andrew and Zhu, Jun-Yan and Ramanan, Deva},
  booktitle={Proceedings of the IEEE/CVF Conference on Computer Vision and Pattern Recognition},
  pages={12882--12891},
  year={2022}
}

@inproceedings{chung2024depth,
  title={Depth-regularized optimization for 3d gaussian splatting in few-shot images},
  author={Chung, Jaeyoung and Oh, Jeongtaek and Lee, Kyoung Mu},
  booktitle={Proceedings of the IEEE/CVF Conference on Computer Vision and Pattern Recognition},
  pages={811--820},
  year={2024}
}

@inproceedings{yu2021pixelnerf,
  title={pixelnerf: Neural radiance fields from one or few images},
  author={Yu, Alex and Ye, Vickie and Tancik, Matthew and Kanazawa, Angjoo},
  booktitle={Proceedings of the IEEE/CVF conference on computer vision and pattern recognition},
  pages={4578--4587},
  year={2021}
}

@inproceedings{mildenhall2019llff,
  title={Local Light Field Fusion: Practical View Synthesis with Prescriptive Sampling Guidelines},
  author={Mildenhall, Ben and Srinivasan, Pratul P. and Ortiz-Cayon, Rodrigo and Kalantari, Nima Khademi and Ramamoorthi, Ravi and Ng, Ren and Kar, Abhishek},
  booktitle={ACM Transactions on Graphics (TOG)},
  volume={38},
  number={4},
  pages={1--14},
  year={2019},
  publisher={ACM}
}

@article{barron2022mipnerf360,
  title={Mip-NeRF 360: Unbounded Anti-Aliased Neural Radiance Fields},
  author={Barron, Jonathan T. and Mildenhall, Ben and Verbin, Dor and Srinivasan, Pratul P. and Hedman, Peter},
  journal={arXiv preprint arXiv:2111.12077},
  year={2022}
}

@article{jensen2014dtu,
  title={Large scale multi-view stereopsis evaluation},
  author={Jensen, Rasmus and Dahl, Anders Lindbjerg and Vogiatzis, George and Tola, Engin and Aanaes, Henrik},
  journal={Proceedings of the IEEE Conference on Computer Vision and Pattern Recognition},
  pages={406--413},
  year={2014}
}

@inproceedings{xu2022sinnerf,
  title={Sinnerf: Training neural radiance fields on complex scenes from a single image},
  author={Xu, Dejia and Jiang, Yifan and Wang, Peihao and Fan, Zhiwen and Shi, Humphrey and Wang, Zhangyang},
  booktitle={European Conference on Computer Vision},
  pages={736--753},
  year={2022},
  organization={Springer}
}

@article{ho2022imagen,
  title={Imagen Video: High Definition Video Generation with Diffusion Models},
  author={Ho, Jonathan and Chan, William and Saharia, Chitwan and Whang, Jay and Gao, Ruiqi and Gritsenko, Alexey and Kingma, Diederik P and Poole, Ben and Norouzi, Mohammad and Fleet, David J and others},
  journal={arXiv preprint arXiv:2210.02303},
  year={2022}
}

@article{guo2023animatediff,
  title={AnimateDiff: Animate Your Personalized Text-to-Image Diffusion Models without Specific Tuning},
  author={Guo, Yuwei and Yang, Ceyuan and Rao, Anyi and Liang, Zhengyang and Wang, Yaohui and Qiao, Yu and Agrawala, Maneesh and Lin, Dahua and Dai, Bo},
  journal={International Conference on Learning Representations},
  year={2024}
}

@article{blattmann2023stable,
  title={Stable Video Diffusion: Scaling Latent Video Diffusion Models to Large Datasets},
  author={Blattmann, Andreas and Dockhorn, Tim and Kulal, Sumith and Mendelevitch, Daniel and Kilian, Maciej and Lorenz, Dominik and Levi, Yam and English, Zion and Voleti, Vikram and Letts, Adam and Jampani, Varun and Rombach, Robin},
  journal={arXiv preprint arXiv:2311.15127},
  year={2023}
}

@article{zhang2023universal,
  title={Universal Guidance for Diffusion Models},
  author={Zhang, Bowen and Li, Chunyuan and Zhu, Yichen and Zhang, Jianfeng and Liu, Jianfeng and Gao, Jianfeng},
  journal={arXiv preprint arXiv:2303.08919},
  year={2023}
}

@inproceedings{yu2023freedom,
  title={FreeDoM: Training-Free Energy-Guided Conditional Diffusion Model},
  author={Yu, Jiwen and Wang, Yinhuai and Zhao, Chen and Ghanem, Bernard and Zhang, Jian},
  booktitle={Proceedings of the IEEE/CVF International Conference on Computer Vision (ICCV)},
  pages={1859--1869},
  year={2023}
}
}


\end{document}